\documentclass[conference]{IEEEtran}
\IEEEoverridecommandlockouts
\usepackage{algorithm}
\usepackage{algpseudocode}
\usepackage{cite}
\usepackage{orcidlink}
\usepackage{hyperref}
\usepackage{amsmath,amssymb,amsfonts}
\usepackage{graphicx}
\usepackage{textcomp}
\usepackage{xcolor}
\usepackage{bm}

\newcommand{\tp}{\textsc{TangentProp} }
\newcommand{\ntp}{$n$-\tp}

\def\BibTeX{{\rm B\kern-.05em{\sc i\kern-.025em b}\kern-.08em
    T\kern-.1667em\lower.7ex\hbox{E}\kern-.125emX}}

\begin{document}
\title{
A Quasilinear Algorithm for Computing Higher-Order Derivatives of Deep Feed-Forward Neural Networks
}

\author{
\IEEEauthorblockN{Kyle R. Chickering\orcidlink{https://orcid.org/0000-0003-4249-0266}}
\IEEEauthorblockA{\textit{UC Davis Dept. Mathematics}}
\IEEEauthorblockA{\textit{krchickering@ucdavis.edu}}
}

\maketitle
\thispagestyle{plain}

\begin{abstract}
    The use of neural networks for solving differential equations is practically difficult due to the exponentially increasing runtime of autodifferentiation when computing high-order derivatives. We propose \ntp, the natural extension of the \tp formalism \cite{simard1991tangent} to arbitrarily many derivatives. \ntp computes the exact derivative $d^n/dx^n f(x)$ in quasilinear, instead of exponential time, for a densely connected, feed-forward neural network $f$ with a smooth, parameter-free activation function. We validate our algorithm empirically across a range of depths, widths, and number of derivatives. We demonstrate that our method is particularly beneficial in the context of physics-informed neural networks where \ntp allows for significantly faster training times than previous methods and has favorable scaling with respect to both model size and loss-function complexity as measured by the number of required derivatives. The code for this paper can be found at \href{https://github.com/kyrochi/n\_tangentprop}{\texttt{https://github.com/kyrochi/n\_tangentprop}}.
\end{abstract}


\section{Introduction}
Physics-informed neural networks (PINNs) were introduced as a numerical method to solve forward and inverse problems involving differential equations using neural networks instead of traditional numerical solvers \cite{raissi2019physics}. Their use has recently come under scrutiny for several reasons, including a lack of high-accuracy results, poor run-times compared to standard numerical methods, and complicated training dynamics \cite{mcgreivy2024weak}. Due to methodological issues in the aforementioned studies, including the failure to use the well-established strategies outlined in \cite{wang2023expert}, we believe that the pursuit of PINNs should not be abandoned and to that end, propose an algorithm which partially addresses the valid concerns over PINN training times.

A primary reason to use PINNs over standard numerical methods is because by approximating the solution to an ODE or PDE by a neural network we obtain a $C^\infty$ approximation to the model solution which can be evaluated at arbitrary points in the domain, as well as allowing for the study of high-order derivatives (see Figures~\ref{fig:profile_3_panel} and~\ref{fig:profile_4_panel}). This strength is also a weakness, since during training we must repeatedly take derivatives of the neural network with respect to the inputs. This is done using autodifferentiation \cite{baydin2018automatic, raissi2019physics}, which suffers from unfavorable scaling when taking multiple derivatives. In particular, taking $n$ derivatives of a neural network $f$ with $M$ parameters gives the exponential runtime $\mathcal{O}(M^n)$. For training PINNs we often need to take two or more derivatives, and in many practical applications this exponential runtime already becomes prohibitive. Furthermore this difficulty cannot simply be overcome by horizontally scaling compute, since repeated applications of autodifferentiation cannot be parallelized on a GPU due to the recursive nature of computing high-order derivatives.

In this paper we introduce \ntp, which addresses these issues by computing $\frac{d^n}{dx^n}f(x)$ in quasilinear $\mathcal{O}(e^{\sqrt{n}}M)$ time instead of the exponential time $\mathcal{O}(\frac{e^{\sqrt{n}}}{n}M^n)$. \ntp is 
an exact method, and thus there is no accuracy degradation when using this method. \ntp is the natural extension of the \tp formalism \cite{simard1991tangent} to $n$ derivatives. \tp was introduced in the context of MNIST digit classification as a way to enforce a smoothness condition on the first derivative of a neural network based classifier. The motivation for \ntp starts from the observation that for PINN training we only need higher-order derivatives with respect to the network inputs, not with respect to the network weights. In practice the dimensionality of the input data $d$ is much smaller than the number of parameters $M$, and therefore it is unnecessary to compute a fully filled out computational graph for all higher order derivatives. Instead of building the full computational graph, we directly differentiate the network during the forward pass and can thus compute all $n$ derivatives with respect to the network inputs during a single forward pass.

Our contributions are three-fold
\begin{enumerate}
    \item We derive the \ntp formalism and give an algorithm for it's implementation.
    \item We show that for simple network architectures consisting of stacked linear, densely connected layers with the $\tanh$ activation function, our method empirically follows the theoretical scaling laws for a variety of widths, depths, and batch sizes.
    \item We show that in the context of PINN training our method can significantly reduce training time and memory requirements when compared to the standard PINN implementation. 
\end{enumerate}
\section{PINN Training}
PINNs are neural networks trained to approximate the solution to a given ODE or PDE \cite{raissi2019physics}. Because neural networks with smooth activation functions are $C^\infty$ function approximators, and $C^\infty$ functions are dense in most function space which are used in practice (like the $L^2$ based Sobolev family of function spaces), we can train a neural network using gradient descent to approximate the solution to a given differential equation. It is this simple observation that led to the introduction of PINNs in the paper \cite{raissi2019physics}. We give a brief overview of the methodology behind PINNs, but refer the reader to the recent surveys \cite{antonion2024machine, farea2024understanding, wang2024recent} and the references therein, as the literature abounds with introductory material on PINNs.

Let $u_\theta(\bm{x})$ be a feed-forward, densely connected, neural network with parameters $\theta$. Our goal is to optimize these parameters in such a way that $u_\theta$ is an approximate $C^\infty$ solution to the differential equation $F(\partial^\alpha u; \bm{x}) = 0$ for some multi-index $\alpha$ \cite{evans2022partial}. We train the neural network on the discrete domain $\Omega = \{\bm{x}_1, \cdots \bm{x}_N\}$ using the loss target
\begin{align}\label{eq:vanilla_loss}
    L(u_\theta) = \frac{1}{N}\sum_{k=1}^N|F(\partial^\alpha u_\theta; \bm{x}_k)|^2 + \text{BC},
\end{align}
which is the mean-squared error (MSE) of the differential equation residual with BC being appropriately enforced boundary conditions. Since $u_\theta$ is $|\alpha|$-times continuously differentiable we can use autodifferentiation to exactly compute the derivatives appearing above, and thus by the approximation theorem \cite{cybenko1989approximation, hornik1991approximation}, if the solution to $F(\partial^\alpha u; \bm{x}) = 0$ lies in a space in which $C^\infty$ functions are dense, we can theoretically train a neural network to approximate the true solution of the PDE to arbitrary precision.

In practice this is not so easy: PINN training is complicated by needing to enforce boundary conditions, which introduces problems inherent to multi-target machine learning \cite{bischof2021multi, wang2020understanding}. There is also the problem of effectively choosing collocation points from the domain, as well as the problem of choosing a good network architecture \cite{wang2023expert, yang2024deeper}. PINN training appears naturally unstable, likely due to a poorly conditioned Hessian, and thus training these networks can be difficult \cite{krishnapriyan2021characterizing, rathore2024challenges, wang2022and}. This instability has further been related to the conditioning of a specific differential operator related to the underlying differential operator \cite{de2023operator}. PINNs also appear to struggle fitting high-frequency components of the target solution \cite{markidis2021old} as a consequence of spectral bias (f-principle) \cite{rahaman2019spectral, xu2019frequency}. 

Additionally, convergence under the loss function~\eqref{eq:vanilla_loss} is often slow. In practice it is usually better to use "Sobolev training" \cite{czarnecki2017sobolev, maddu2022inverse, son2021sobolev} which replaces~\eqref{eq:vanilla_loss} with the Sobolev-norm \cite{evans2022partial} loss function
\begin{align}\label{eq:sobolev_loss}
    L_{\text{Sobolev}}^{(m)}(u_\theta) = \frac{1}{N}\sum_{k=1}^N\sum_{j=0}^m\,Q_j|\nabla_{\bm{x}}^{j}F(\partial^\alpha u_\theta; \bm{x}_k)|^2 + \text{BC},
\end{align}
where $Q_j$ are relative weights which add additional hyperparameters to the training \cite{maddu2022inverse}. While this loss function generally improves accuracy, it also requires computing $m$ extra derivatives of the neural network $u_\theta$. Due to the nature of autodifferentiation, this trade-off quickly becomes costly and in practice we can often only train with $m=1$ or $m=2$, despite the fact that higher $m$ usually results in better solution accuracy. \ntp makes this trade-off much cheaper and we hope that future authors are able to train with $m=4$ or higher while retaining reasonable training times.

The appearance of high-order derivatives is also not uncommon in PINN applications. Wang et al. \cite{wang2023asymptotic} show that to satisfactorily compute successive high-order unstable shock profiles for the Burgers, De Gregorio, and Boussinesq equations, one must take high-order derivatives. For example, to compute the $m$-th smooth, self-similar shock profile for Burgers equation one must take $2m + 3$ derivatives. The authors compute the first and second profiles which already requires taking five derivatives and is already slow. Using \ntp we are able to compute the third and fourth profiles in this paper in a reasonable amount of time (See section~\ref{sec:ss_burg} below).

To our knowledge there has been no prior work which directly addresses the exponential runtime of autodifferentiation in the literature. Instead, works aimed at making training more computationally efficient rely on augmenting the PINN training with some sort of numerical differentiation \cite{chiu2022can, sharma2022accelerated}, pre-training or transfer learning methods \cite{jahani2024enhancing, penwarden2023unified}, or efficient sampling of the collocation points \cite{nabian2021efficient}. There are too many articles in this direction to compile an exhaustive list, and we instead refer again to the many surveys and studies which abound \cite{antonion2024machine, farea2024understanding, wang2024recent, wang2023expert}. We stress that \ntp is an exact method and all of the aforementioned studies would be accelerated by adopting our proposed methodology.
\section{\ntp}
\subsection{Autodifferentiation}
Autodifferentiation is a method for computing the derivatives of a neural-network using the network's computational graph \cite{baydin2018automatic, paszke2019pytorch}. Autodifferentiation is usually applied in the context of gradient descent for optimizing neural networks where it is used to efficiently and exactly compute the first partial derivatives of the loss with respect to each of the network weights \cite{paszke2019pytorch}. It is usually applied once per training iteration, and outputs a vector of first-order partial derivatives with respect to all network inputs (including the weights). It is able to do this computation in $\mathcal{O}(M + d)=\mathcal{O}(M)$ time, where $M$ is the number of model parameters and $d$ is the dimensionality of the input.

It is less common for autodifferentiation to be applied repeatedly. However this repeated application is the key to the effectiveness of PINNs \cite{raissi2019physics}. Because there are $\mathcal{O}(M^2)$ second-order partial derivatives, and $\mathcal{O}(M^n)$ $n$-th order partial derivatives, the repeated application of auto-differentiation takes $\mathcal{O}(M^n)$ to compute all $n$-derivatives. This bound does not fully account for the amount of time required to take the $n$-th derivative of the activation function, see below. The full runtime is $\mathcal{O}\left(\frac{e^{\sqrt{n}}}{n}M^n\right)$. For small networks and a low number of derivatives this runtime is acceptable, but the exponential growth makes training large PINN models or PINN models for equations involving high-order derivatives prohibitively difficult. Furthermore, as discussed above, it appears beneficial to use the Sobolev loss~\eqref{eq:sobolev_loss} which requires taking even more derivatives. In practice PINN training becomes prohibitively slow when computing more than three or four derivatives of the network.

It is easy to see why autodifferentiation is not the right tool for computing derivatives in the PINN loss function: we do not need every partial derivative computed by autodifferentiation. In fact, we only need the $n$-th partial derivatives corresponding to the network inputs, i.e. 
\begin{align*}
    \nabla_{\bm{x}}^n f (\bm{x}) := \frac{\partial^n}{\partial {\bm{x}}^n}f(\bm{x}),
\end{align*}
which is often a sparse subset of the total derivatives computed by auto-differentiation. In what follows we propose \ntp, which is an efficient quasilinear algorithm to compute only these partial derivatives that are needed for the PINN loss function.

\subsection{Tangent Prop and \ntp}\label{sec:tp_ntp}
The \tp formalism introduced in \cite{simard1991tangent} derives the exact forward and backward propagation formulas for the derivative of a deep feed-forward network. This was done in the context of MNIST digit classifictation based on the observation that the classifier (the neural network) should be invariant to rotations of the input digits, and thus the derivative of the classifier with respect to the inputs (the tangent vector) should be zero. This effectively enforces a first-order (derivative) constraint on the classifier.

Let $\sigma$ be an activation function, $\bm{a}^\ell$ be the activations at the $\ell$-th layer, $\bm{w}$ be the weight matrix, and $\bm{x}^0$ be the network inputs, the authors in \cite{simard1991tangent} derive the formula for the first derivative $\bm{\gamma}$ of the network using a single forward pass
\begin{subequations}
\begin{align}
    a_i^\ell &= \sum_{j}w_{ij}^\ell x_j^{\ell - 1}, \qquad x_j^{\ell - 1} = \sigma(a_{j}^{\ell-1}),\label{eq:forward_act} \\
    \gamma_i^\ell &= \sum_{j}w_{ij}^\ell \xi_j^{\ell - 1}, \qquad \xi_j^{\ell - 1} = \sigma'(a_j^\ell)\gamma_j^\ell.\label{eq:tangent_1_prop}
\end{align}
\end{subequations}
The formula~\eqref{eq:tangent_1_prop} is derived by applying the chain rule to the per-layer activation formula~\eqref{eq:forward_act}.

We can naturally extend \tp to compute $n$ derivatives in a single forward pass by applying Fa\`a di Bruno's formula \cite{roman1980formula}, which generalizes the chain rule to multiple derivatives. Fa\`a di Bruno's formula states that for the composition of $C^n$ continuous functions $f$ and $g$ we have
\begin{align}\label{eq:faa}
    (f(g(x))^{(n)} = \sum_{\bm{p}\in \mathcal{P}(n)}\,C_{\bm{p}} \left(f^{|\bm{p}|}\right)(g(x))\prod_{j=1}^{n}\left(g^{(j)}(x)\right)^{p_j},
\end{align}
where the sum is taken over the set $\mathcal{P}(n)$ of partition numbers of order $n$, which consists of all tuples $\bm{p}$ of length $n$ and satisfying $\sum_{j=1}^n j p_j = n$, $0 \leqslant p_j \leqslant n$, and $|\bm{p}|=\sum_{j} p_j$. The constants $C_{\bm{p}}$ are explicitly computable and the size of the set $\mathcal{P}$ is found using the partition function $p(n) = |\mathcal{P}(n)|$ whose combinatorial properties are well-studied \cite{brualdi2010}.

Since neural networks with smooth activation functions are $C^\infty$, we can apply our new formalism, which we call \ntp, to take arbitrarily many derivatives of deep feed-forward neural-networks. Using Fa\`a di Bruno's formula~\eqref{eq:faa} in place of the chain rule in~\eqref{eq:forward_act} allows us to compute an arbitrary derivative in the same forward pass that we compute the activations. The formula for the $n$-th derivative $\bm{\gamma}^{(n)}$ is given by
\begin{subequations}\label{eq:ntp}
\begin{align}
    (\gamma^{(n)})_i^\ell &= \sum_{j}\,w_{ij}^\ell (\xi^{(n)})_j^{\ell - 1}, \\ (\xi^{(n)})_j^{\ell - 1} &= \sum_{\bm{p}}C_{\bm{p}}\sigma^{(|\bm{p}|)}(a_j^\ell)\prod_{m=1}^{n}\left((\xi^{(m)})_j^\ell\right)^{p_m}. \label{eq:ntp_forward}
\end{align}
\end{subequations}
where $a_j^\ell$ are the forward activations computed in~\eqref{eq:forward_act}.
Thus, in a single forward pass through the model we can compute all of the required derivatives at once with runtime of $\mathcal{O}(np(n)M)$, where $p(n)$ is the partition function. Thus, we have reduced the exponential runtime from auto-differentiation to a quasilinear runtime (see Algorithm~\ref{alg:ntp} and the tighter bound derived below). Note that the derivatives must be computed in order, since $\bm{\xi}^{(m)}$ depends on $\bm{\xi}^{(k)}$ for all $k < m$.

The coefficients $C_{\bm{p}}$ appearing in~\eqref{eq:ntp_forward} are the coefficients of the Bell polynomials of the second kind (See \cite{roman1980formula} and references therein). These are well-studied and explicitly computable, and for efficiently implementing Algorithm~\ref{alg:ntp} we recommend pre-computing and caching the required coefficients (see our implementation code for more details).

\begin{algorithm}\label{alg:ntp}
\caption{Forward Pass with Higher-Order Derivatives}
\begin{algorithmic}[1]
\Procedure{Forward}{$x, n$} \Comment{$x$: input, $n$: derivative order}
\If{$n = 0$}
\For{$layer$ in $\textsc{Network}$}
\State $x \gets layer(\sigma(x))$ \Comment{$\sigma$ is activation function}
\EndFor
\State \Return $x$
\EndIf
\State $y \gets \text{array of length } n + 1$
\State $y_0 \gets L_1(x)$ \Comment{$L_1$ is first layer}
\State $y_1 \gets L_1(\mathbf{1}) - b_1$ 
\For{$i \gets 2$ to $n$}
\State $y_i \gets L_1(\mathbf{0}) - b_1$ 
\EndFor
\For{$L$ in $\textsc{Network}[2:]$}
\State $a \gets \sigma(y_0, n)$ 
\For{$i \gets n$ down to $1$}
\State $z \gets \mathbf{0}$
\For{$(c, e)$ in $\textsc{Bell}[i]$} 
\State $s \gets \sum e$
\State $t \gets \prod_{j: e_j \neq 0} y_j^{e_j}$
\State $z \gets z + c \cdot t \cdot a_s$
\EndFor
\State $y_i \gets z$
\EndFor
\State $y_0 \gets L(a_0)$
\For{$i \gets 1$ to $n$}
    \State $y_i \gets L(y_i) - b$ \Comment{Subtract bias}
\EndFor
\EndFor
\State \Return $y$
\EndProcedure
\end{algorithmic}
\end{algorithm}

For the sake of completeness we give a tighter bound on the runtime which takes into account the dependence on $n$ of the summation appearing in~\eqref{eq:faa}. The combinatorial properties of the summation over the integer partitions $\mathcal{P}(n)$ are well studied. In particular, the partition function $p(n)$ counts the number of integer partitions and thus $p(n)=|\mathcal{P}(n)|$. A well-known and classical result of Hardy and Ramanujan \cite{hardy1917asymptotic} provides an upper and lower bound on the partition function $p(n)$ which then yields the asymptotic behavior
\begin{align*}
    p(n) = \mathcal{O}\left(
    \frac{e^{\sqrt{n}}}{n}\right).
\end{align*}
This implies the more refined runtime depending on $n$ and $M$ of $\mathcal{O}(n p(n)M) \sim \mathcal{O}\left(e^{\sqrt{n}} M\right)$ which is our claimed quasilinear bound. Note that during autodifferentiation the Fa\`a di Bruno formula must implicitly be applied to the activation function $\sigma$ and therefore autodifferentiation has an actual asymptotic runtime of $\mathcal{O}\left(\frac{e^{\sqrt{n}}}{n}M^n\right)$. Finally we remark that the memory complexity of \ntp is linear at $\mathcal{O}(nM)$ (without a modification from $p(n)$) while the memory complexity of autodifferentiation is exponential at $\mathcal{O}(M^n)$. Thus, not only does our algorithm compute derivatives faster than autodifferentiation, but we can compute more derivatives on the same hardware than is even possible using autodifferentiation.
\section{Experiments}
We begin by demonstrating that for a wide range of feed-forward neural network architectures that our proposed method indeed follows the theoretical asymptotic scaling laws. In particular we consider a standard feed-forward network with uniform width across the layers and the $\tanh$ activation function. Then we use our method to train a PINN model to compute the smooth stable and unstable profiles for the self-similar Burgers profile using the methodology proposed in \cite{wang2023asymptotic}. This problem requires taking a large number of derivatives for the training to converge, and we demonstrate that our method is able to break through the computational bottlenecks imposed by autodifferentation and we are able to compute higher-order profiles which were previously either impractical or impossible to compute using autodifferentiation on a single GPU.

\subsection{Implementation Details and Methodology}
We run our experiments using Python and PyTorch \cite{paszke2019pytorch}. In particular we implement \ntp as a custom forward method for a PyTorch \texttt{torch.nn.Module} implementation of a deep feed-forward network. For the PINN experiments we then build a custom PINN training framework to handle the PINN training loop, and we use an open-source L-BFGS implementation \cite{shi2020lbfgs} instead of the PyTorch L-BFGS due to the latter not supporting line-search\footnote{See for example \url{https://discuss.pytorch.org/t/optimizer-with-line-search/19465} and \url{https://discuss.pytorch.org/t/l-bfgs-b-and-line-search-methods-for-l-bfgs/674} for further discussion of this deficiency.}. All experiments were run locally on a single NVIDIA A6000 GPU.

\subsection{Forward-Backward Pass Times}
PyTorch implements several asynchronous optimizations that make benchmarking difficult. To mitigate the effects of built-in optimizations on our benchmarking we implement the following steps
\begin{enumerate}
    \item Randomly shuffle the experiments over all parameters to ensure that execution order is not a factor in the results.
    \item Synchronize CUDA between runs.
    \item Enable \texttt{cudnn.benchmark}.
    \item Run the Python garbage collector between runs.
    \item Use the Python performance counter instead of the timing module.
\end{enumerate}
These mitigation strategies allow for a fairer comparison between autodifferentiation and \ntp.

We first explore the effect of \ntp on the computation of a single forward and backward pass to verify that the empirical performance aligns with the predicted theoretical performance suggested by our derivation. In particular we would expect to see exponential run-times for autodifferentiation and quasilinear run-times for \ntp.

For a given network we compute and time the forward pass through the network, compute the loss outside of a timing module, then compute and time a backwards pass through the network. The total time includes the time it takes to compute the loss function, while the forward and backwards pass times only include the time it takes to compute the given pass.

\begin{figure}[h]
    \centering
    \includegraphics[width=0.95\linewidth]{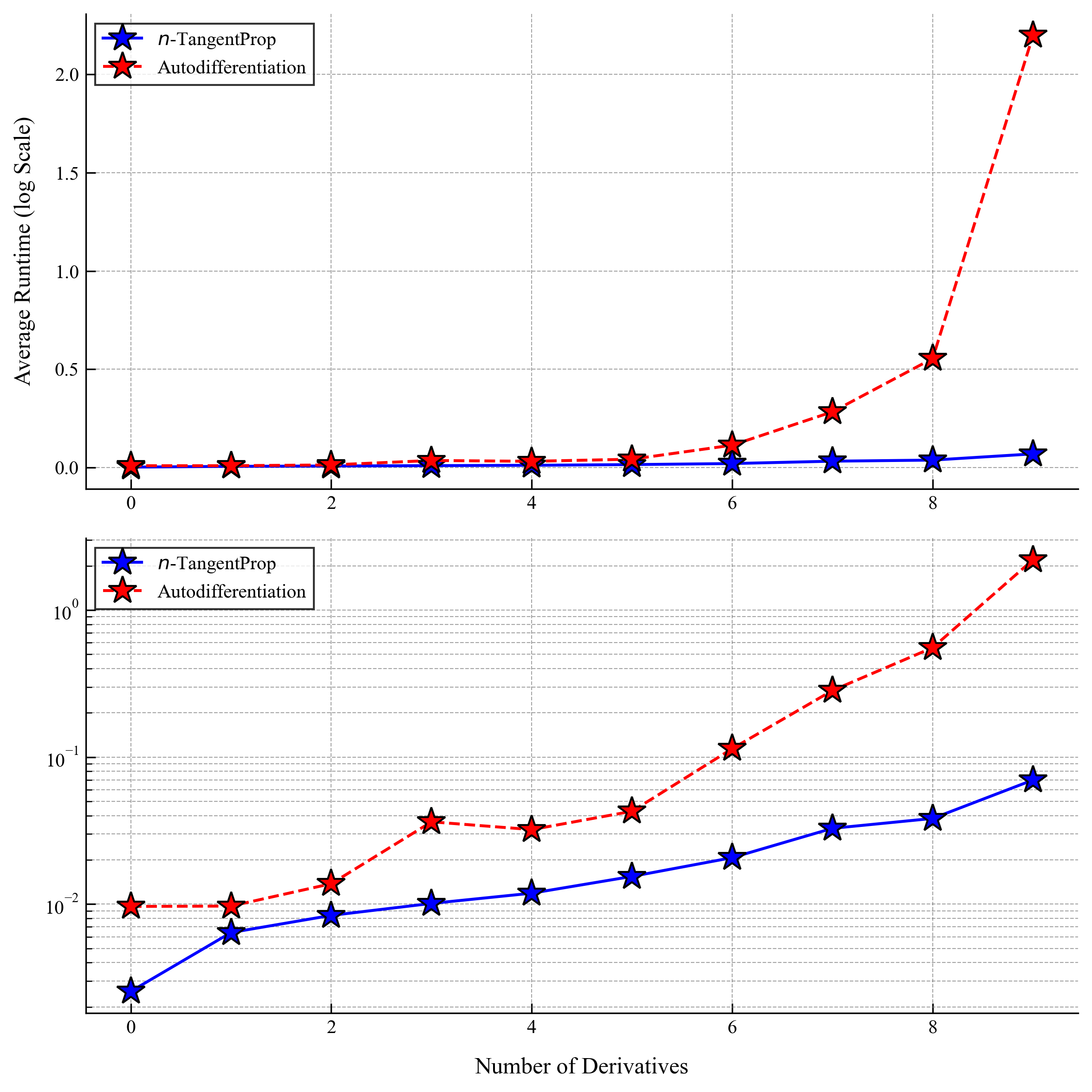}
    \caption{Average runtime for a combined forward and backwards pass using autodifferentiation (red) and $n$-TangentProp (blue). The top and bottom frames show the same data, however the bottom frame is plotted with a logarithmic $y$-axis. Each model is run 100 times and the average for each trial is plotted. The network has 3 hidden layers of 24 neurons each, a common PINN architecture. The batch size is $2^{8}=256$ samples. The forward and backwards pass times are shown separately in Figures~\ref{fig:forward_times} and~\ref{fig:backward_times} respectively.}
    \label{fig:forward_pass_times_const_params}
\end{figure}

\begin{figure}
    \centering
    \includegraphics[width=0.95\linewidth]{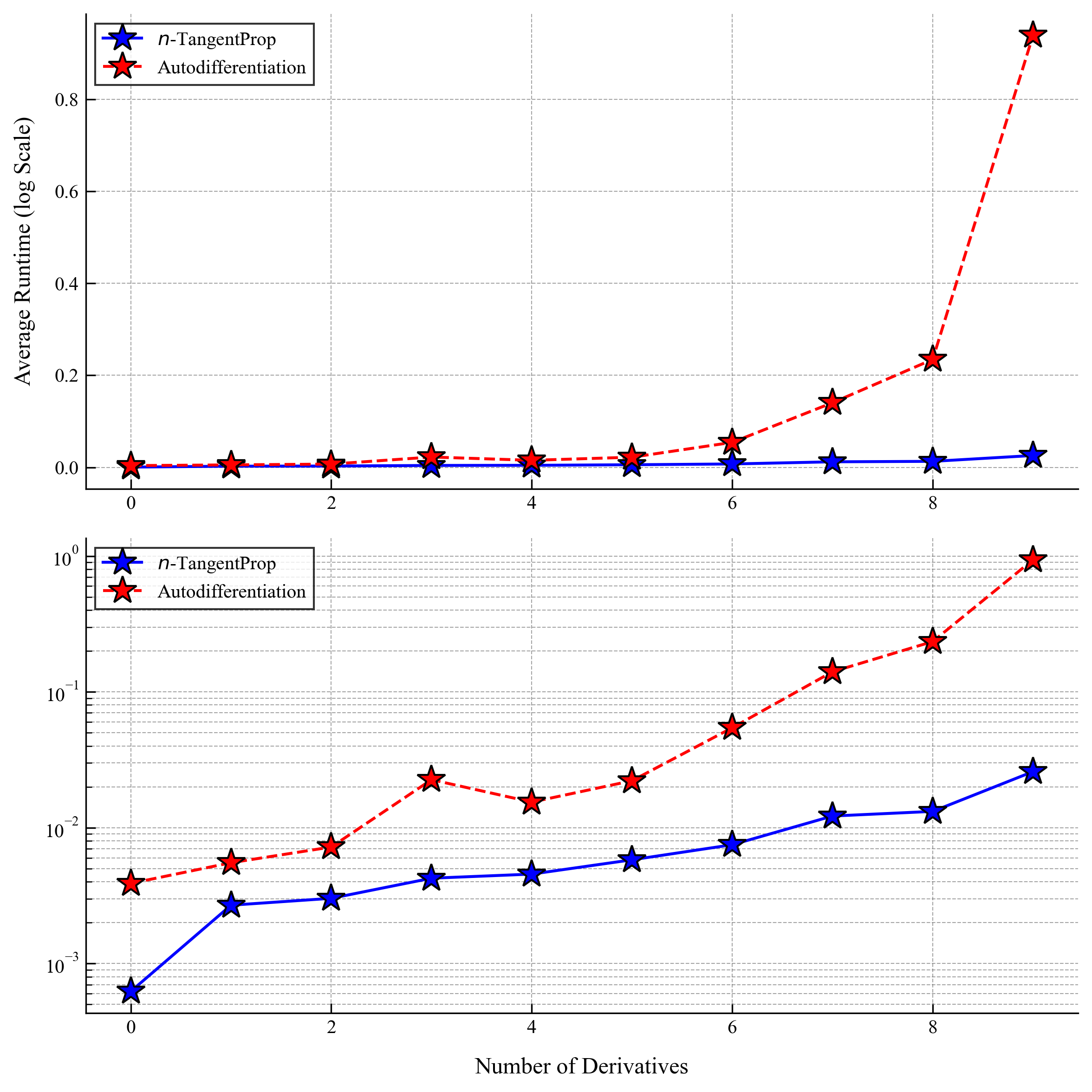}
    \caption{Forward pass times for the model shown in Figure~\ref{fig:forward_pass_times_const_params}. The top and bottom frames show the same data, however the bottom frame is plotted with a logarithmic $y$-axis. Each model is run 100 times and the average for each trial is plotted. The network has 3 hidden layers of 24 neurons each, a common PINN architecture. The batch size is $2^{8}=256$ samples.}
    \label{fig:forward_times}
\end{figure}

\begin{figure}
    \centering
    \includegraphics[width=0.95\linewidth]{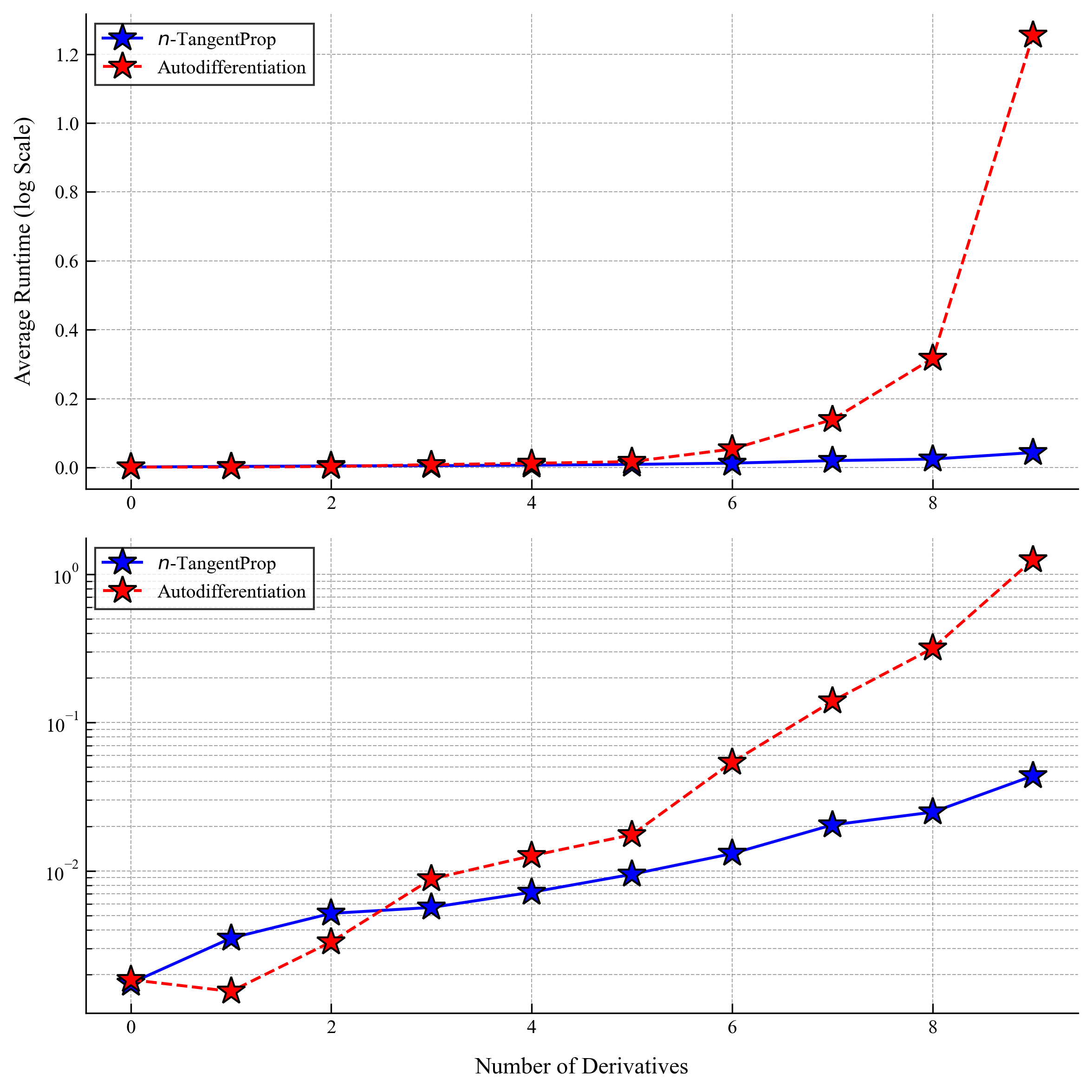}
    \caption{Backwards pass times for the model shown in Figure~\ref{fig:forward_pass_times_const_params}. The top and bottom frames show the same data, however the bottom frame is plotted with a logarithmic $y$-axis. Each model is run 100 times and the average for each trial is plotted. The network has 3 hidden layers of 24 neurons each, a common PINN architecture. The batch size is $2^{8}=256$ samples.}
    \label{fig:backward_times}
\end{figure}

For a fixed network size, we find that the end-to-end times for a combined forward and backward pass for autodifferentiation scales exponentially and that \ntp scales roughly quasilinearly (Figure~\ref{fig:forward_pass_times_const_params}), keeping with the theoretical predictions made above by our formalism. We can further decompose this total execution time into its forward and backward times (Figure~\ref{fig:forward_times} and Figure~\ref{fig:backward_times} respectively). We see that the \ntp formalism gives more significant performance gains during the forward pass when compared to the backwards pass. We hypothesize that this is due to PyTorch graph optimizations that are applied automatically to the autodifferentiation implementation and are not included in our \ntp implementation. This difference is seen most plainly in Figure~\ref{fig:backward_times}, where autodifferentiation outperforms \ntp in backwards pass times for small numbers of derivatives. 

\begin{figure}[h]
    \centering
    \includegraphics[width=0.95\linewidth]{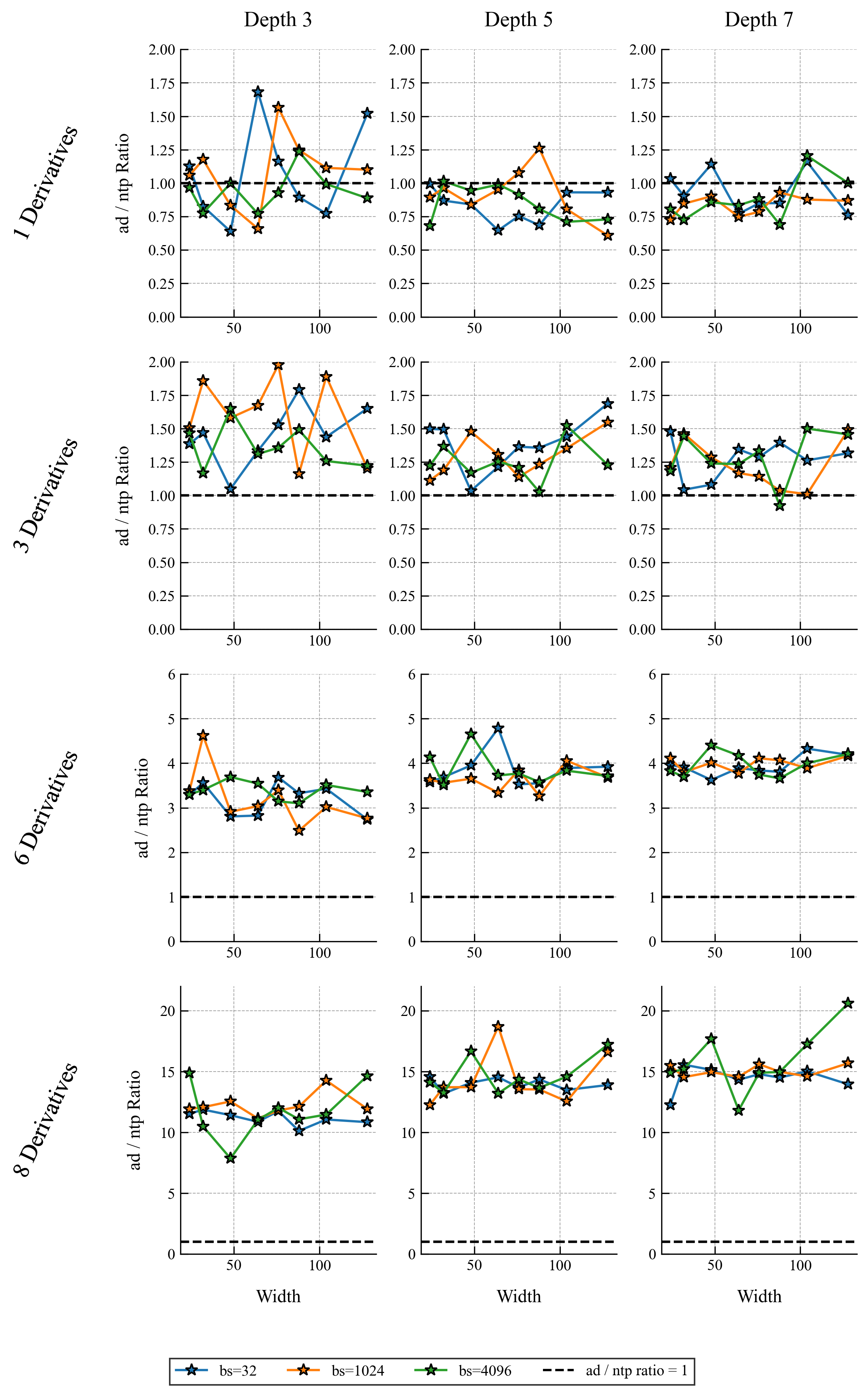}
    \caption{The ratio of forward pass run times between autodifferentiation and \ntp for a variety of network architectures, input batch sizes, and number of derivatives. A ratio greater than $1$ indicates that \ntp was faster than autodifferentiation. The baseline ratio of $1$ is plotted as a horizontal dashed line. All plotted data points represent the average of $100$ trials.}
    \label{fig:ftime_grid}
\end{figure}

\begin{figure}[h]
    \centering
    \includegraphics[width=0.95\linewidth]{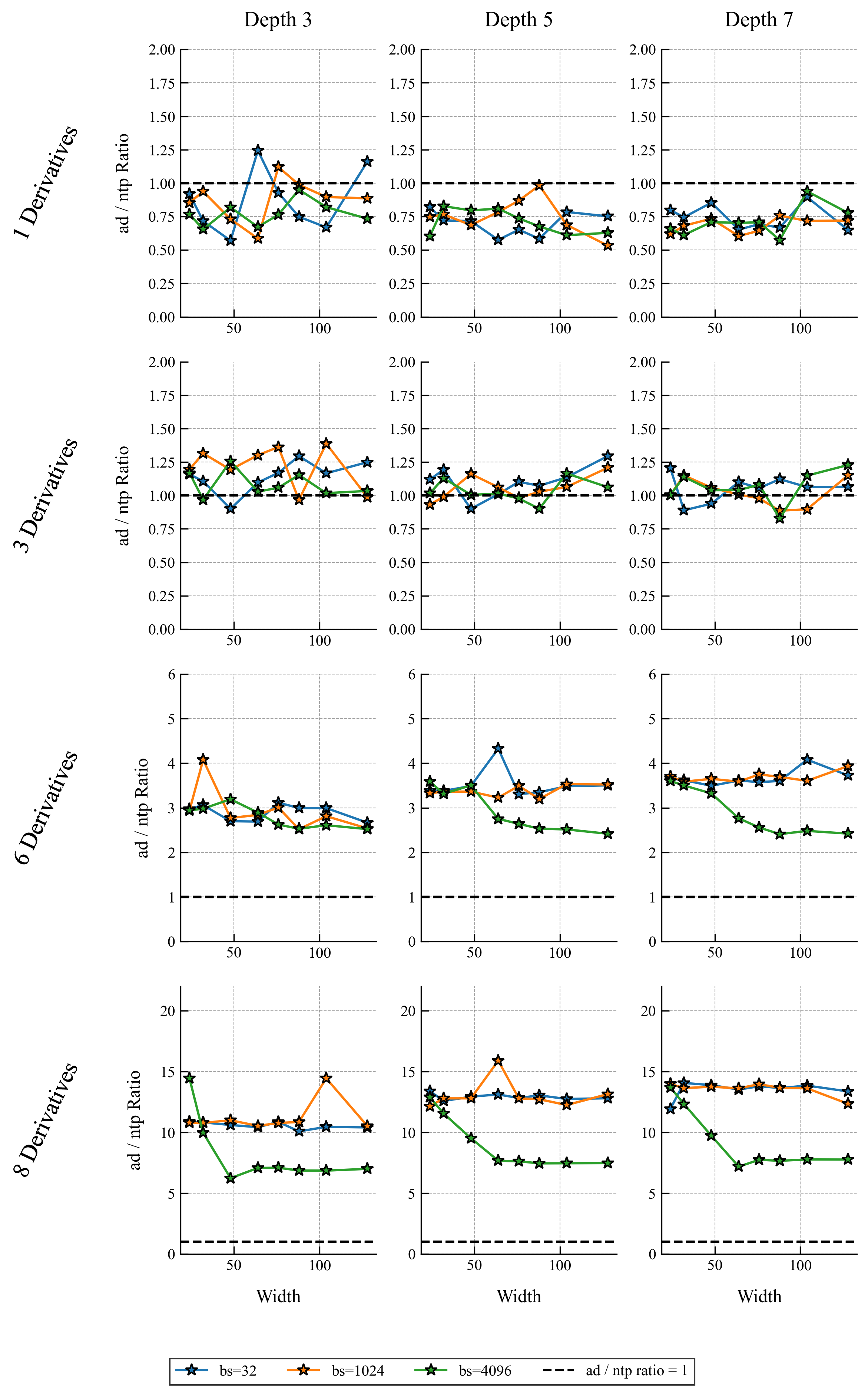}
    \caption{The ratio of combined forward-backward pass run times between autodifferentiation and \ntp for a variety of network architectures, input batch sizes, and number of derivatives. A ratio greater than $1$ indicates that \ntp was faster than autodifferentiation. The baseline ratio of $1$ is plotted as a horizontal dashed line. All plotted data points represent the average of $100$ trials. The forward pass time ratio alone is plotted in Figure~\ref{fig:ftime_grid}.}
    \label{fig:ttime_grid}
\end{figure}

We run extensive experiments to analyze the effect of varying batch size, network width, network depth, and number of derivatives. The results of the forward passes are summarized in Figure~\ref{fig:ftime_grid}, and the results of the combined forward and backward pass are summarized in Figure~\ref{fig:ttime_grid}.

We point out several salient features in these results. First, we observe a performance gap between \ntp and autodifferentiation for low derivatives. This is likely a consequence of implementation details, rather than a deficiency with the proposed methodology. The PyTorch implementation of autodifferentiation is heavily optimized for execution on a GPU, and while our implementation makes attempts at closing this performance gap, it is written in Python, rather than a lower level language such as \texttt{C++}, and lacks sophisticated optimization strategies. We suspect that a more refined implementation would close the gap seen for low derivatives.

Second, we observe that the performance gains afforded by \ntp decrease as we increase the batch size. We hypothesize that this effect is also due to a lack of optimization to take full advantage of the parallelized computational ability of the GPU in our implementation. For example, our implementation does not fully vectorize the computation of Equation~\ref{eq:ntp_forward} and thus does not take full advantage of the hardware scaling afforded by our GPU. Similarly, we observe that the performance gains from \ntp decrease as we increase the network width. We suspect that this is also a consequence of the lack of vectorization. Increasing either width or batch size scales the compute horizontally, and we have not fully optimized our implementation to account for this horizontally scaled compute.

Third, we observe that for all of the tested derivatives and batch-sizes, the standard PINN architecture of three hidden layers and 24 neuron widths \cite{raissi2019physics} performs better with \ntp than autodifferentiation, at least for derivatives of order three or higher. This suggests that for PINN problems involving higher-order derivatives, \ntp can be used as a drop in replacement without any further implementation tuning (See Section~\ref{sec:pinn_training} below).

Finally, we observe an apparent asymptote for the combined forward-backward pass times as the number of parameters and batch size increase (see the bottom rows of Figure~\ref{fig:ttime_grid}). We suspect that this is because as we increase the relative number of FLOPs the theoretical gains from \ntp begin to dominate the superior optimization in PyTorch's autodifferentiation implementation. We hypothesize that with a stronger \ntp implementation we would begin to see similar asymptotic behaviors even when taking fewer derivatives.

We add that we could not compute more than nine derivatives using autodifferentiation because the required memory exceeded the 49 GB of memory available on our GPU. 

\subsection{End-to-End PINN Training}\label{sec:pinn_training}
Forward-Backward pass times should correlate to end-to-end model training but it is still important to measure the effect of the proposed modifications over the long time-horizons and multiple optimizers present in end-to-end PINN training. For example, our proposed \ntp method uses a different memory footprint for forward pass than autodifferentiation does. It can be difficult to reason theoretically about the effect that such changes will have to the end-to-end performance of machine learning models, and as such, empirical analysis is imperative to rule out performance degradation which arises as a consequence of the complicated end-to-end training testbed.

We find that the widespread use of the L-BFGS optimizer adds to the improvements afforded by \ntp. L-BFGS performs quasi-second order optimization using a line-search \cite{nocedal1999numerical} which requires performing multiple forward passes through the network but only a single backwards pass. Thus, the forward pass performance seen in Figure~\ref{fig:ftime_grid} is expected to dominate and we expect that our unoptimized \ntp algorithm will outperform standard PINN implementations using autodifferentiation.

\subsubsection{Unstable Self-Similar Burgers Profiles}\label{sec:ss_burg}
Burgers equation is the canonical model for 1D shock formation phenomenon \cite{haberman1987elementary} and is given by the PDE
\begin{align}\label{eq:burgers_pde}
    \partial_t u + u\partial_x u = 0.
\end{align}
Due to the nonlinear steepening of the wave profiles leading ultimately to a gradient blowup, studying the behavior of shock solutions near the singularity is difficult. Work in the 20th century explored the use of self-similarity to study the breakdown of smooth solutions near a shock \cite{barenblatt1972self}. Under the self-similar coordinate transformation
\begin{align*}
    u(x, t) = (1-t)^{\lambda}U(x (1-t)^{-1 - \lambda}), \quad X = x (1-t)^{-1 - \lambda},
\end{align*}
the PDE~\eqref{eq:burgers_pde} becomes the ODE \cite{eggers2015singularities} for $U$ in $X$
\begin{align}\label{eq:self_similar_burg}
    -\lambda U + \left((1 + \lambda)X + U\right)U' = 0,
\end{align}
for a scalar valued parameter $\lambda \in \bm{R}^{> 0}$. While the solution of this particular problem is elementary and given implicitly by
\begin{align}\label{eq:burgers_implicit}
    X = -U - CU^{1 + \frac{1}{\lambda}},
\end{align}
the techniques used for the numerical analysis and solution of this problem can be applied to more challenging problems to yield highly non-trivial results about shock formation in complicated nonlinear equations \cite{wang2023asymptotic}.

From~\eqref{eq:burgers_implicit} we observe that the solution $U$ will be smooth whenever $1 + \frac{1}{\lambda}$ is an integer, and be physically realizable (odd) whenever $1+\frac{1}{\lambda} = 2k$ for some positive integer $k$ \cite{eggers2015singularities}. Thus, the possible values of $\lambda$ corresponding to smooth solutions are $\lambda = \frac{1}{2k}$ for $k=1, 2, \cdots$. For all other values of $\lambda$ the solution $U$ will suffer a discontinuity at the origin in one of it's higher-order derivatives.

Our goal for this problem is to find these physically realizable solutions, a problem which is complicated by the fact that the profiles corresponding to $k=2, 3, \cdots$ are physically and numerically unstable \cite{eggers2015singularities}. Traditional solvers will not converge to these solutions, and they do not manifest as real-world shocks due to a collapse towards the solution corresponding to $k=1$. Regardless, these unstable profiles are important to understand mathematically and can give insights into the underlying behavior of certain systems. See for example the papers \cite{chickering2023asymptotically, oh2024gradient}, which apply self-similar methodology to perturbations of the self-similar Burgers equation~\ref{eq:self_similar_burg}.

Wang et al. \cite{wang2023asymptotic} propose a methodology for solving Equation~\ref{eq:self_similar_burg} for an unknown value of $\lambda$ using PINNs to perform a combined forward-inverse procedure and simultaneously solve for $U$ and $\lambda$. Such a methodology demonstrates the advantage that PINNs have in certain numerical settings, since solving this problem using traditional solvers is challenging. We refer the reader to the study by Biasi \cite{biasi2021self} which addresses a similar problem using traditional numerical methods and highlights the attendant difficulties.

The primary observation in \cite{wang2023asymptotic} is that a solution to~\eqref{eq:self_similar_burg} is smooth for all values of $\lambda$, except at the origin, where a discontinuity will appear for derivatives $n \geqslant 1 + \frac{1}{\lambda}$. Thus, if we restrict the value of $\lambda$ to $[1/3, 1]$ and enforce a smoothness condition on the third derivative of our neural network, we will converge to the unique smooth profile (if one exists) in the range $[1/3, 1]$. This is because for any non-smooth profile in this range, the third derivative or lower \textbf{must} be non-smooth at the origin. To find higher-order smooth profiles we can look between $[1/(2k+3), 1/(2k+1)]$ and enforce a smoothness condition on the $2k+3$-th derivative.

Using a PINN, we can enforce differentiability at the origin by taking sufficiently many derivatives there, since the neural network solution is smooth. This forces the solution to be smooth, which in turn gives gradient signal to push $\lambda$ towards a valid value. We loosely follow the training schedule used in \cite{wang2023asymptotic} to compute the first, second, third, and fourth profiles in quasilinear, rather than exponential time. The authors in \cite{wang2023asymptotic} only computed the first and second profiles, so our results represent the first time the third and fourth profiles have been computed using PINNs. We stress that computing these solutions requires taking many derivatives and thus the \ntp formalism is well-suited for this type of problem. Additionally, computing the third or higher profile is extremely computationally intensive.

We note in passing that we were unable to reproduce the accuracy results claimed by the Wang et al. paper \cite{wang2023asymptotic} and the authors do not provide access to their code. Regardless, our work is orthogonal to theirs and any future attempts at a reproduction of their work will benefit from using \ntp.

We run our self-similar Burgers experiments using 64-bit floating point precision. We give a detailed description of our methodology and results below in Appendix~\ref{app:burgers}, which we hope will contribute to the reproducibility of the \cite{wang2023asymptotic} results.

\begin{figure}[h]
    \centering
    \includegraphics[width=0.95\linewidth]{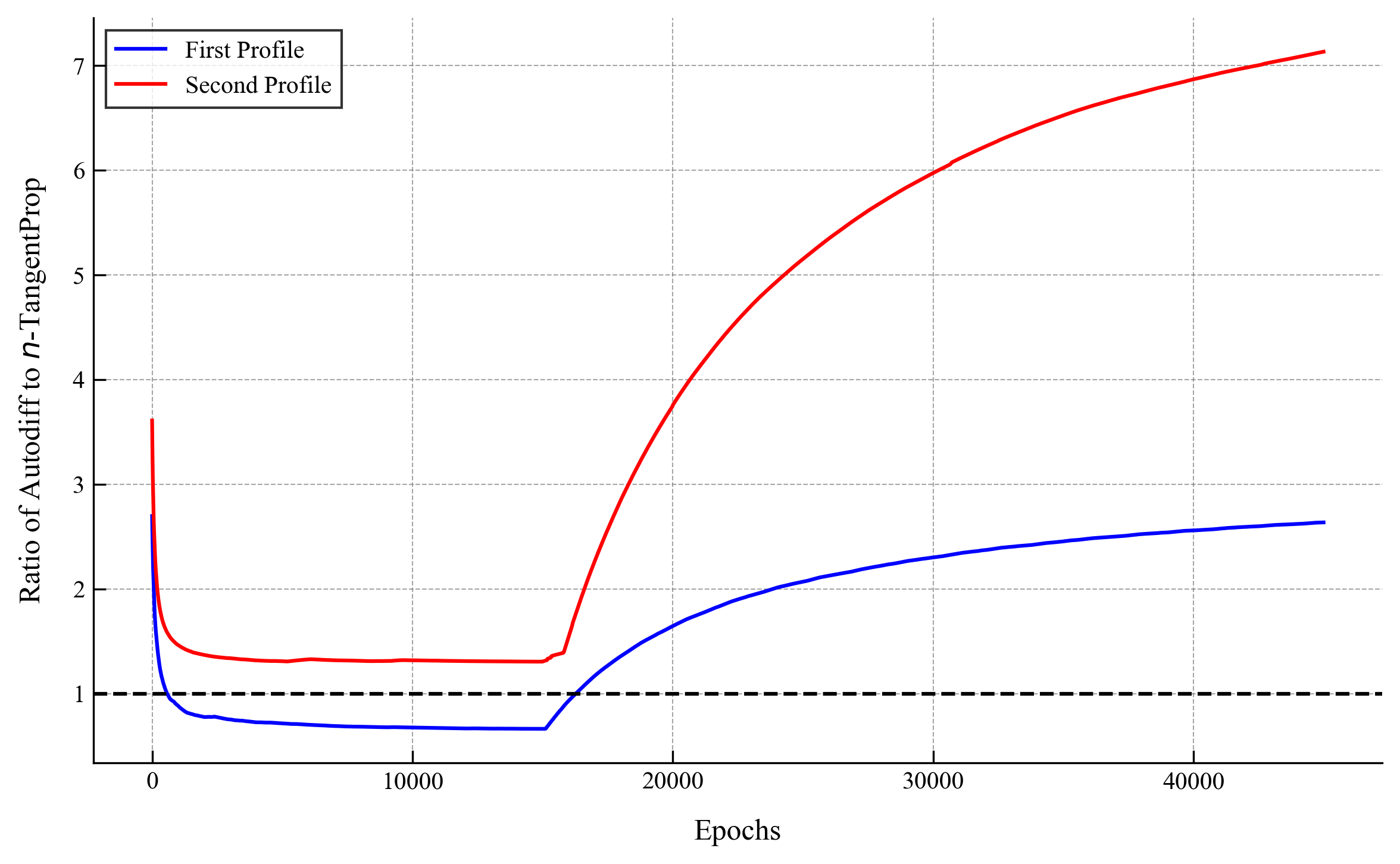}
    \caption{Results from training a PINN to find the first smooth profile for Equation~\ref{eq:self_similar_burg}. The model is trained for 15k epochs using the Adam optimizer and 30k epochs using L-BFGS. The top panel reports our training losses, the middle panel reports $\lambda$ as a function of epochs, and the bottom panel shows the ratio in runtime between autodifferentiation and \ntp as a function of number of epochs. The bottom pannel shows that \ntp is 2.5x faster for end-to-end training than autodifferentiation. The horizontal line in the bottom panel indicates a runtime ratio of 1.}
    \label{fig:profile_1}
\end{figure}

Figure~\ref{fig:profile_1} plots the ratio of execution times of autodifferentiation over \ntp and
shows that we obtain significant speedups in computation time using \ntp instead of autodifferentiation.  We were only able to compute the timing comparison between autodifferentiation and \ntp for the first two profiles since the computational time for the third profile using autodifferentiation exceeded our allowable computation time of 24 hours. For the first profile, which requires taking three derivatives, we obtain an end-to-end speed up of over 2.5 times. For the second profile, which requires taking five derivatives, we obtain an end-to-end speed up of over 7 times. We were able to compute the third profile, which requires taking seven derivatives, in a little under 1 hour using \ntp, and the projected time for auto-differentiation was over 25 hours, giving an expected speed-up of at least 25 times.

Using \ntp we were also able to compute the fourth profile, which requires taking nine derivatives. Using \ntp we were able to run the 45k epochs in a little under an hour and a half. We discuss our results further in Appendix~\ref{app:burgers}, which we think are interesting in their own right. We stress that computing this fourth profile is untenable using autodifferentiation, as the time and space complexity render attempts at computation impossible. We estimate that computing the fourth profile using autodifferentiation would take at least 100 hours (about four days).

Observe from Figure~\ref{fig:profile_1} that the most dramatic improvements come when we switch to the L-BFGS optimizer, which uses multiple forward passes to perform a line search. Because \ntp has more favorable forward pass dynamics (c.f. Figures~\ref{fig:forward_times} and~\ref{fig:backward_times}), the performance improvements are much more pronounced during L-BFGS optimization. This emphasizes the advantage afforded by \ntp: to obtain high-accuracy results we often use L-BFGS and Sobolev loss (see Equation~\ref{eq:sobolev_loss}). These two accuracy improvements require taking higher-order derivatives more frequently, which are the two areas that \ntp shows the best improvement in. Thus, for the high-accuracy training phase for PINNs, \ntp yields significant performance improvements.

We suspect that the dip below a ratio of 1 that we see in Figure~\ref{fig:profile_1} for computing the first profile can be mitigated through further optimizations of our implementation, and we hypothesize once again that the dip is likely due to efficiencies afforded by graph pruning and operator fusing in PyTorch.

\begin{figure}[h]
    \centering
    \includegraphics[width=0.95\linewidth]{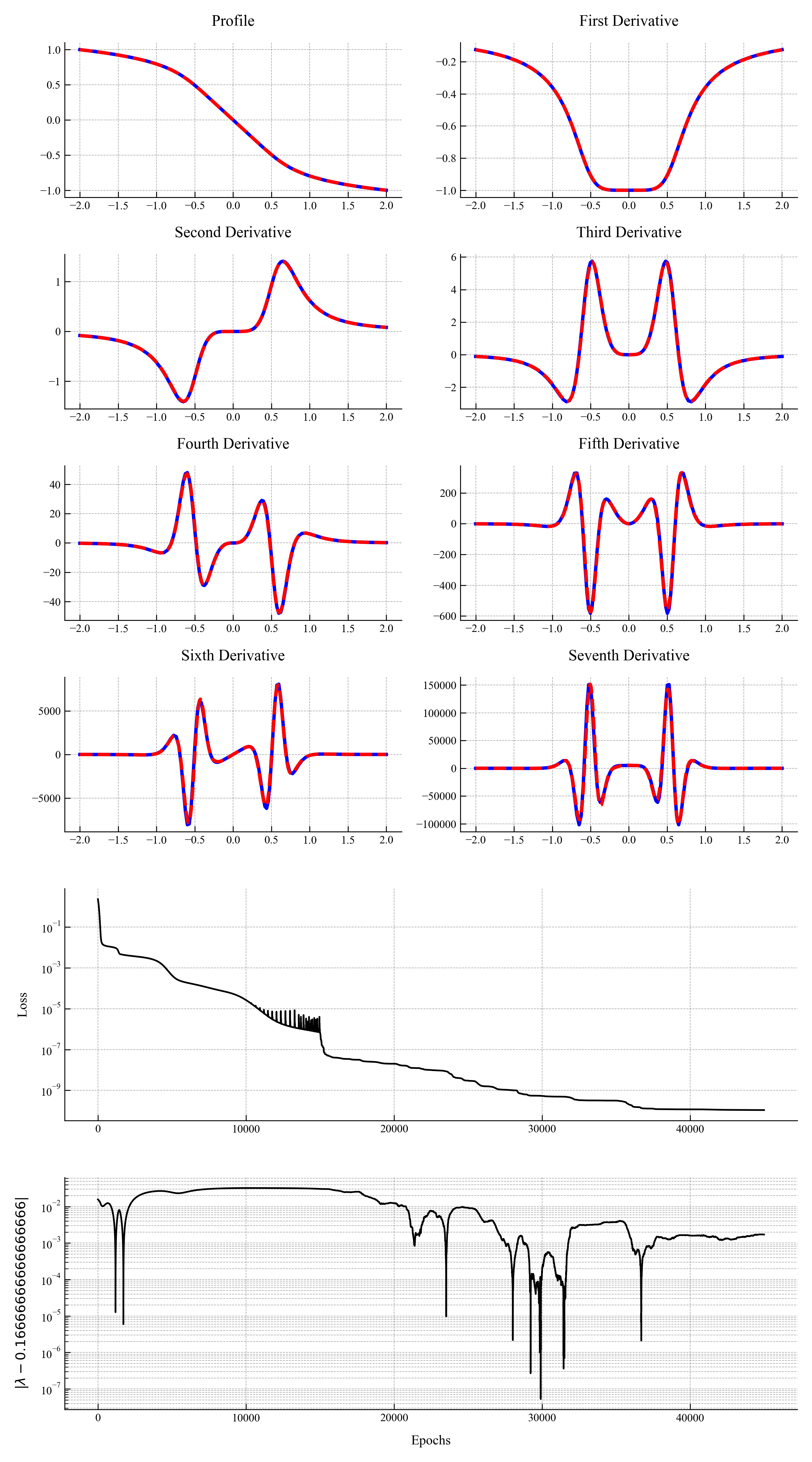}
    \caption{Results from training a PINN to solve~\eqref{eq:self_similar_burg} with $\lambda$ constrained to the range $[1/7, 1/5]$. The only smooth solution to~\eqref{eq:self_similar_burg} contained in this parameter range corresponds to $\lambda=1/6$. The first four rows show our learned solution (dashed red) and its derivatives compared to the true solution (solid blue). The second to last row shows the PINN training loss as a function of epochs. The model was trained for 15k epochs using the Adam optimizer and 30k epochs using the L-BFGS optimizer. The bottom row shows the inferred value for the parameter $\lambda$ as a function of epochs. The bottom two rows are plotted with a logarithmic $y$-axis.}
    \label{fig:profile_3_panel}
\end{figure}

Figure~\ref{fig:profile_3_panel} shows the result of training a PINN to find the third smooth profile of~\eqref{eq:self_similar_burg}. This is the first time that we are aware of that this profile has been numerically computed with a floating value of $\lambda$ using either a PINN or a traditional solver. We show the learned solution with a dashed red line superimposed over the true solution in blue. This profile is already computationally expensive to compute using autodifferentation and \ntp opens the door to performing novel studies on equations requiring a high-number of derivatives.
\section{Conclusion}
We introduced the \ntp formalism and demonstrated both theoretically and empirically that implementing our formalism in the context of PINNs dramatically reduces training times. We showed that for derivative-intensive PINN applications like finding high-order solutions to self-similar equations, \ntp not only offers improvements in end-to-end training times but allows the computation of previously untenable solutions. Our results are a step in the direction of making PINNs a more competitive numerical method for difficult forward and inverse problems. We recommend that our formalism be adopted by PINN implementations going forward to ensure faster training of PINNs.

We hope that our work allows the PINN community to explore more complicated problems, deeper and wider network architectures, and allow for researchers who do not have access to powerful computer to participate in furthering PINN research.

In this paper we have not focused on the optimization of our algorithm. We think that with optimization choices like implementing the underlying logic in \texttt{C++} instead of Python that the performance gap between \ntp and autodifferentiation would widen even further.

\section*{Acknowledgment}
The author thanks Dr. Patrice Simard for stimulating discussions throughout the completion of this work. This work was partially completed while the author was employed by Hummingbird.ai.

\bibliographystyle{siam}
\bibliography{bib}

\appendix
\subsection{Additional Details for the Self-Similar Burgers Experiments}\label{app:burgers}
We report the results from running the self-similar Burgers experiment to find the smooth stable and unstable profiles. We were not able to reproduce the accuracy reported in \cite{wang2023asymptotic}, however we think that our results are important in demonstrating that their proposed methodology is robust, at least in theory. Furthermore, we report several new observations that we think are relevant to the training dynamics for such a problem.

We train our network using a Sobolev loss function (see Equation~\ref{eq:sobolev_loss} function with $m=1$ and additionally add a high-order loss term
\begin{align*}
    L^*(u_\theta) = \frac{1}{N^*}\sum_{k=1}^{N^*} |\partial_x^n R(u_\theta, x_k)|^2,
\end{align*}
where $R$ is the residual of the self-similar Burgers equation and the samples $x_k$ are taken from a small subset of collocation points centered at the origin, not the entire training domain. Our implementation contains many more subtle details that we omit for the sake of brevity and we encourage the reader to download our code to see the full implementation.

\begin{figure}[h]
    \centering
    \includegraphics[width=0.95\linewidth]{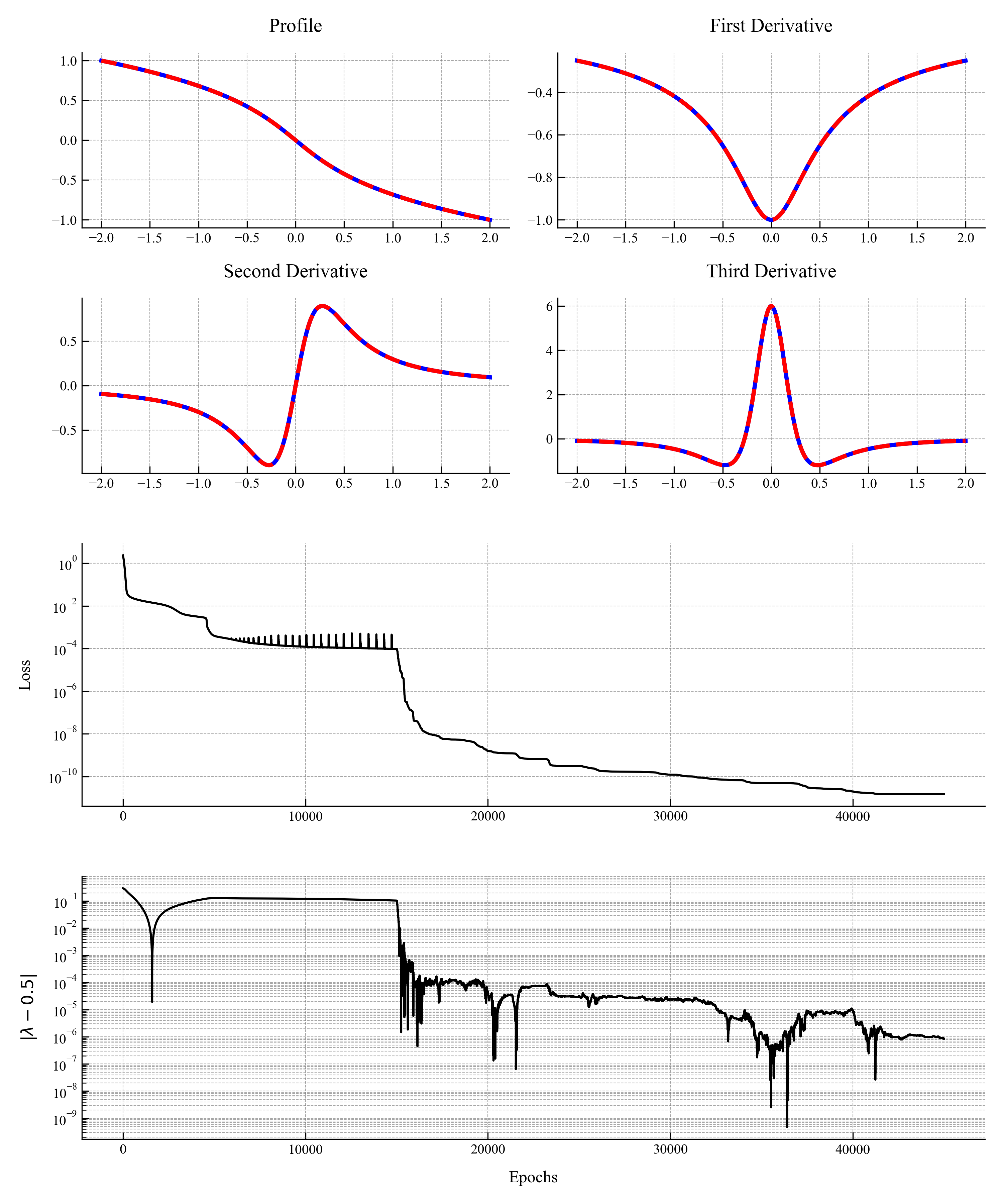}
    \caption{Results from training a PINN to solve~\eqref{eq:self_similar_burg} with $\lambda$ constrained to the range $[1/3, 1]$. The only smooth solution to~\eqref{eq:self_similar_burg} contained in this parameter range corresponds to $\lambda=1/2$. The first two rows show our learned solution (dashed red) and its derivatives compared to the true solution (solid blue). The second to last row shows the PINN training loss as a function of epochs. The model was trained for 15k epochs using the Adam optimizer and 30k epochs using the L-BFGS optimizer. The bottom row shows the inferred value for the parameter $\lambda$ as a function of epochs. The bottom two rows are plotted with a logarithmic $y$-axis.}
    \label{fig:profile_1_panel}
\end{figure}

While we were not able to match the accuracy reported by \cite{wang2023asymptotic}, we were able to get our implementation to perform well in finding the first three smooth solutions to~\eqref{eq:self_similar_burg}. However our method did not satisfactorily solve for the fourth profile. We discuss this in more detail below. 

\begin{figure}[h]
    \centering
    \includegraphics[width=0.95\linewidth]{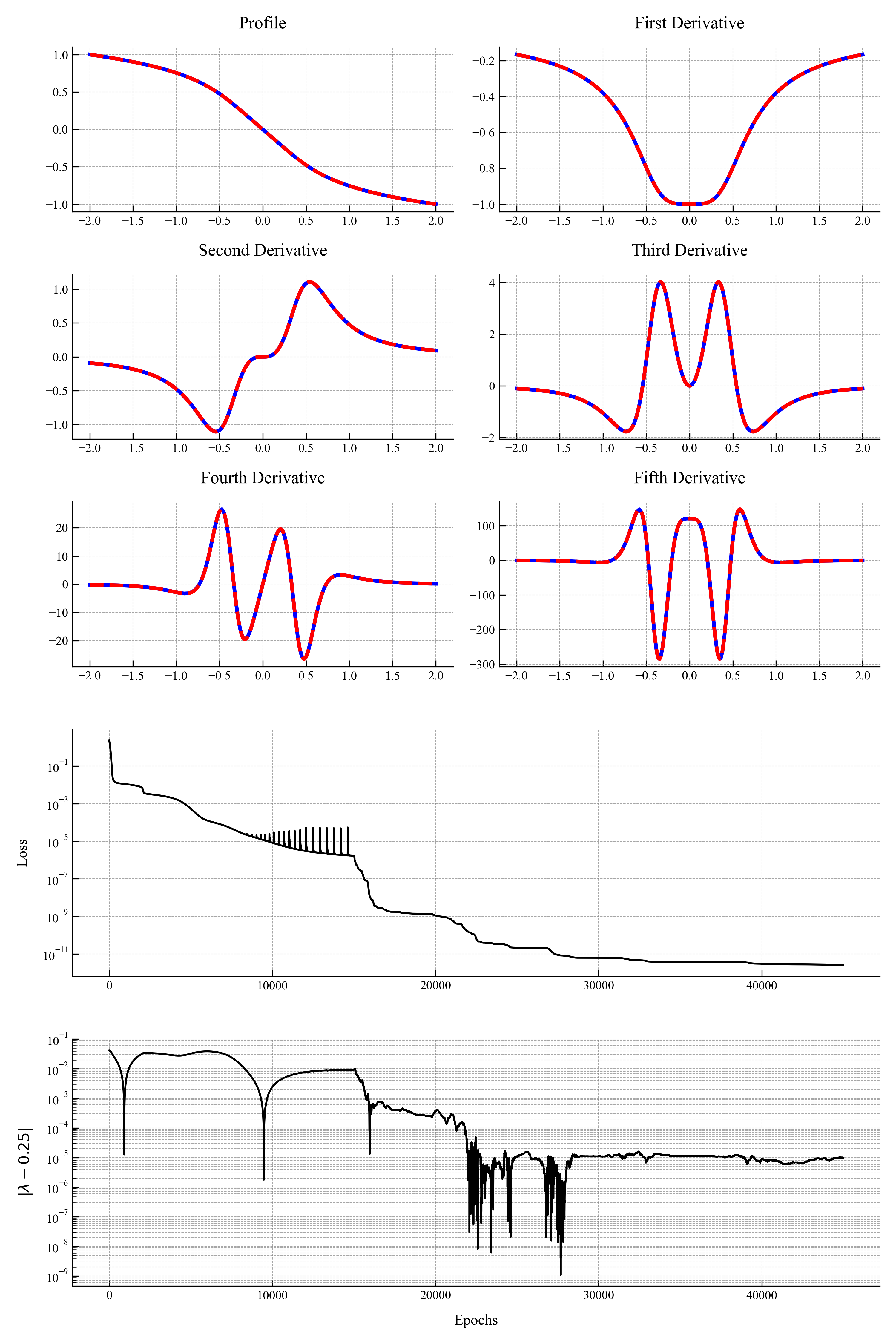}
    \caption{Results from training a PINN to solve~\eqref{eq:self_similar_burg} with $\lambda$ constrained to the range $[1/5, 1/3]$. The only smooth solution to~\eqref{eq:self_similar_burg} contained in this parameter range corresponds to $\lambda=1/4$. The first three rows show our learned solution (dashed red) and its derivatives compared to the true solution (solid blue). The second to last row shows the PINN training loss as a function of epochs. The model was trained for 15k epochs using the Adam optimizer and 30k epochs using the L-BFGS optimizer. The bottom row shows the inferred value for the parameter $\lambda$ as a function of epochs. The bottom two rows are plotted with a logarithmic $y$-axis.}
    \label{fig:profile_2_panel}
\end{figure}

For all of the profiles we found, we report our inferred value of $\lambda$ as a function of training epochs. We think that this metric is an important scalar quantity and it's evolution is not reported in the original paper \cite{wang2023asymptotic}. Of particular importance is the apparent inability of the Adam optimizer to satisfactorily converge to the correct value of $\lambda$. We see a sharp decrease in the error of $\lambda$ once we begin to use the L-BFGS optimizer. We think that this phenomenon is interesting and may indicate that the first order derivatives of the residual with respect to the parameter $\lambda$ is insufficient to capture the true dependency of the solution on $\lambda$. Understanding this dependency more deeply may lead to better training algorithms for these types of problems.

\begin{figure}[h]
    \centering
    \includegraphics[width=0.95\linewidth]{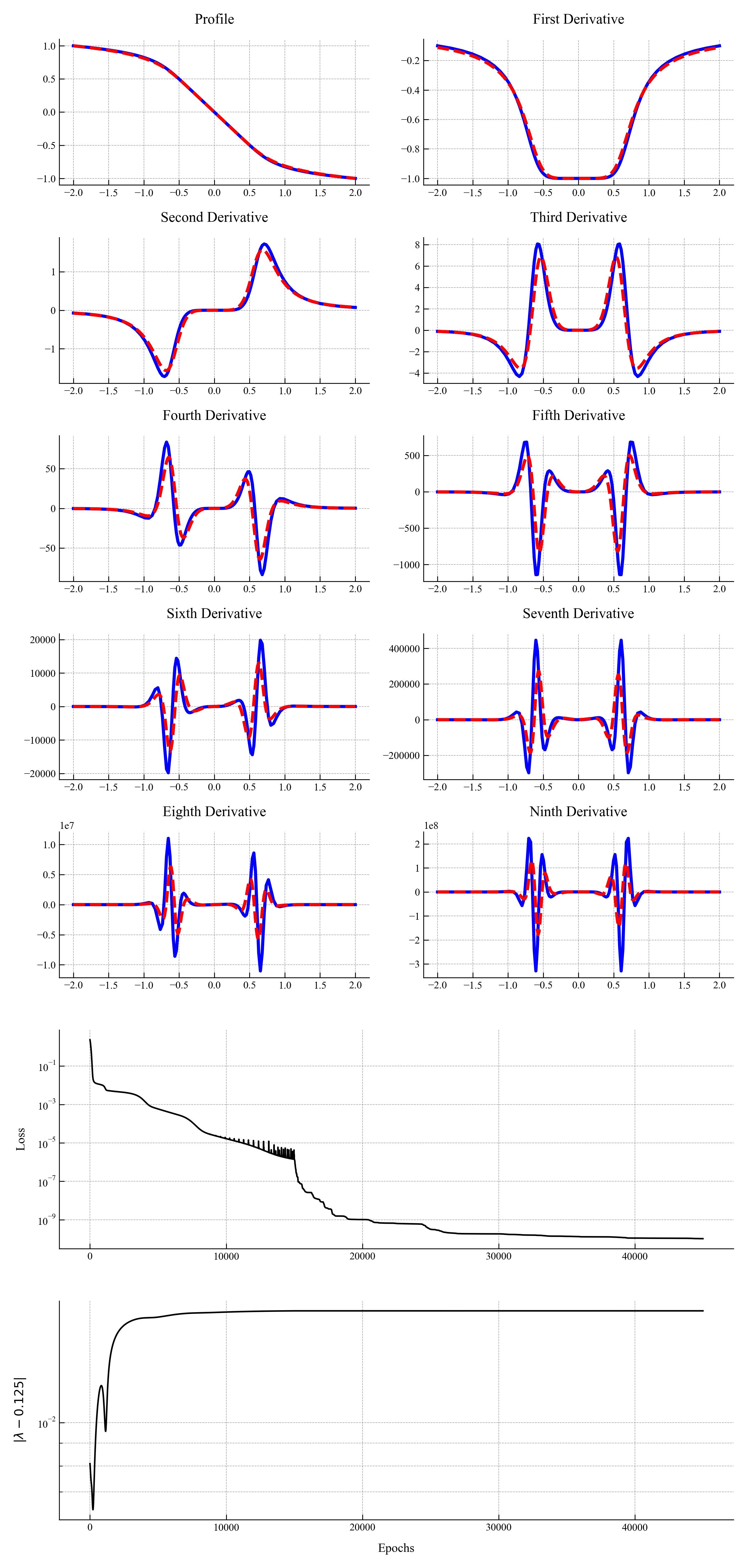}
    \caption{Results from training a PINN to solve~\eqref{eq:self_similar_burg} with $\lambda$ constrained to the range $[1/9, 1/7]$. The only smooth solution to~\eqref{eq:self_similar_burg} contained in this parameter range corresponds to $\lambda=1/8$. The first five rows show our learned solution (dashed red) and its derivatives compared to the true solution (solid blue). The second to last row shows the PINN training loss as a function of epochs. The model was trained for 15k epochs using the Adam optimizer and 30k epochs using the L-BFGS optimizer. The bottom row shows the inferred value for the parameter $\lambda$ as a function of epochs. The bottom two rows are plotted with a logarithmic $y$-axis.}
    \label{fig:profile_4_panel}
\end{figure}

Notably, our code failed to adequately converge to the fourth profile corresponding to $\lambda=\frac{1}{8}$ (see Figure~\ref{fig:profile_4_panel}). Due to the nature of this work we did not pursue this point further and want to emphasize that we are not claiming that the methodology proposed in \cite{wang2023asymptotic} cannot be applied to higher-order profiles. We hypothesize that the nature of the problem makes it more difficult for PINN or non-PINN solvers to find a solution. We are constraining the ninth derivative to be close to zero near the origin, but due to the relatively large magnitude of the ninth derivative, minor fluctuations in the network output will result in large changes to the ninth derivative. This alone may be enough to render our solver incapable of converging to the desired solution as we are using a fixed ratio to balance the relative terms in our loss function (see \cite{bischof2021multi} for further discussion about why multi-target training is difficult). Put another way, we suspect that the loss function we are using does not properly account for the fact that we are taking nine derivatives and that as a consequence the parameter $\lambda$ is not receiving good gradient signal.

The purpose of this study was not to find the optimal hyperparameters for computing these higher-order profiles. Rather, the point of our study is to demonstrate that computing these higher-order profiles is feasible. We stress that using autodifferentitation to find this fourth profile would likely take several days on a state-of-the-art GPU, and we were able to compute it in less than two hours. We leave the refinement of model accuracy to future studies.

\end{document}